\documentclass[12pt]{elsarticle}

\usepackage{graphicx}
\usepackage{amssymb}
\usepackage{subfig}

\journal{Journal Name}

\begin{document}

\begin{frontmatter}


\title{A Time-Segmented Consortium Blockchain for Robotic Event Registration}
\fntext[]{Supported by the Tezos Foundation through a grant for project RobotChain.}



\author{Miguel Fernandes}
\ead{ivo.fernandes@ubi.pt}
\author{Lu\' is A. Alexandre}
\ead{luis.alexandre@ubi.pt}
\address{Universidade da Beira Interior and Instituto de Telecomunica\c c\~oes\\Covilh\~a, Portugal \\}

\begin{abstract}
A blockchain, during its lifetime, records large amounts of data, that in a common usage its kept on its entirety. In a robotics environment, the old information is useful for human evaluation, or oracles interfacing with the blockchain but it is not useful for the robots that require only current information in order to continue their work. This causes a storage problem in blockchain nodes that have limited storage capacity, such as in the case of nodes attached to robots that are usually built around embedded solutions. This paper presents a time-segmentation solution for devices with limited storage capacity, integrated in a particular robot-directed blockchain called RobotChain. Results are presented regarding the proposed solution that show that the goal of restricting each node's capacity is reached without compromising all the benefits that arise from the use of blockchains in these contexts, and on the contrary, it allows for cheap nodes to use this blockchain, reduces storage costs and allows faster deployment of new nodes.
\end{abstract}

\begin{keyword}
blockchain \sep robotics \sep Tezos \sep time segmentation


\end{keyword}

\end{frontmatter}

%\linenumbers

\section{Introduction}
\label{S:1}

Blockchain technology enables the creation of an immutable electronic ledger of information in a distributed way, and usually leads to innumerable entries over time that new nodes joining the network need to obtain in order to contribute to it. Over time it may amount to a data capacity that embedded systems usually do not have, and, as such, this paper proposes a time-segmented blockchain.
%\color{black}
%This idea allows blockchains with a smaller resource footprint, improving performance and allowing lower costs for the nodes operating in a blockchain network, lower costs in storage and an improved new node deployment time.
This idea allows cheap blockchain nodes with small resource footprints to still be able to join the network. It also improves the performance, lowers storage costs and supports faster new node deployment.
% ue to the lower amount of information needed to obtain, by needing the current segment of the blockchain instead of the entire blockchain, still maintaining the possibility of asynchronously to the work in the blockchain get cold storage data.
%\color{black}
The node may be configured to be either a compute device node, or a cold storage node, where a compute device node will only have the current data segment, and the cold storage node will work with all the segments belonging to the blockchain.
Our proposal uses the open-source Tezos blockchain and adapts it to our requirements.

Each segment of the network is connected to the previous segment by having the hash of the last block of the previous segment stored in the second block of the new segment and taking advantage of the protocol activation feature present on the Tezos blockchain. This way, the blockchain maintains its integrity through its life span.
% \color{black}
With this method it is possible to use a faster but limited memory, such as a RAM disk, or cheaper storage methods, improving the speed and decreasing the requirements for the deployment of the blockchain.
% \color{black}z
%, while using the main storage device as a cold storage.

%The time segmentation process is a initialisation parameter with either the block level or with a time stamp where the segmentation should happen. 
%The segmentation process will be aligned with the blocks per cycle parameter and the time between blocks parameter.

% quando se seguemtnta a rede, quem fica com os segmentos? 
%
%
\section{Related Work}
%TODO: AS IS FROM MIT PAPER

There are already several proposals to integrate the blockchain technology with robots, but none that covers either RobotChain use cases or segmentation techniques.

%\color{black}
Ferrer \cite{Ferrer2016} presents blockchain technology as a mean to improve robot swarms by solving problems, such as data confidentiality, distributed decision making, dynamic environment working capacity without master control program modification and a mean for legal responsibility for the members of the swarm. This last feature allows "integration" of robotic swarms with the human society. 
%Ferrer \cite{Ferrer2016} presents how blockchain can improve robotic swarm systems by solving some existing issues. Those issues are data confidentiality, distributed decision making, ability to work in different and dynamic environments without changes to the control program and a way to ensure safety and legal responsibility for the robotic nodes in the swarm in order to be \lq integrated\rq\ with the human society.
%It is also referred that by adding blockchain in the concept of swarm robotics, it is a deviation of the minimalism of the common research practices used in swarm robotics.

%\subsection{RoboChain: A Secure Data-Sharing Framework for Human-Robot Interaction}

In \cite{Strobel2018} a proof-of-concept method is presented using blockchain smart contracts. These smart contracts allow improved security of the robotic swarm, improved the stability of the swarm coordination mechanisms and adds a way of expelling rogue members from the swarm.
This concept has been studied for its performance in decision making regards of presence or absence of Byzantine robots.
Byzantine Fault tolerance is the concern for fault-tolerance in distributed computer systems where components may fail, be unreliable, or be rogue agents.
%\subsection{Towards blockchain-based robonomics: autonomous agents behavior validation}

%In \cite{Danilov2018} a model for a kind of trading marked named \textit{robonomics} is presented. 
Danilov\cite{Danilov2018} presents a trading model market named \textit{robonomics}.
It is focused on agent-based systems, where the behaviour is described as non-deterministic finite state automata, presenting a Model Checking verification technique in order to detect and filter malfunctioning agents.
The validation method has the possibility to be implements on a consensus algorithm or as part of a blockchain application. % can be implemented on a consensus protocol or part of a blockchain decentralised application.
Duckietown platform, with moving robots that follow a defined set of instructions, was used as a real life test.
%As a real live test, a prototype implementation of Duckietown with moving robots is provided, following a set of instructions related to movement.
%\subsection{The blockchain: a new framework for robotic swarm systems}

In Ferrer \textit{et. al.} \cite{Ferrer2018},  RoboChain is presented. It's a learning framework that attempts to solve various issues related to information sharing.
These issues are privacy of personal information and machine learning data sharing, that allow multiple robots to work in different places while sharing their knowledge.
%, a learning framework, called \textit{RoboChain}, is presented. It attempts to solve privacy issues related to using personal information with blockchain technology, sharing data and machine learning models, allowing multiple robotic units to work at different places, sharing their data and their knowledge.
%It uses the latest technologies related to blockchains and machine learning in embedded devices such as low-cost robotic units.
This approach assumes a consortium blockchain where there is no public access and a level of trust between said parties is assumed. Since it is a consortium blockchain, the presence of a "malicious entity" is not considered, but it still provides a way to verify the integrity of the interactions and learning in the blockchain.
%https://en.wikipedia.org/wiki/Byzantine_fault_tolerance

%\subsection{Chained of Things: A Secure and Dependable Design of Autonomous Vehicle Services}
%\color{black}
In \cite{8377911}, a blockchain framework for a secure ride-sharing service between autonomous vehicles and passengers is described. It uses a  blockchain as a communication mechanism that is dependable and trustworthy.

In the white paper by Chorus\cite{chorus2018}, a blockchain operating as a network of vehicles under 5G networks is presented. It raises feasibility concerns regarding transaction processing and smart contract usage. It uses the Tezos network as a basis due to its Turing complete smart contract and built-in formal verification of the programming language, allowing high security for the blockchain and its participants.
This allows focusing in the creation of the mobile ad hoc blockchains that allow disconnections from the main network and allowing sub-networks to perform transactions.
It also presents smart contracts and  "orchestration and choreography protocols that facilitate, verify and enact with computing means a negotiated agreement between consenting parties". It presents the concept of vehicles self ownership, exchanging services, like transportation for parking space or battery charging.
It also presents a set of goals for this system to be successful, such as scalability, cost efficiency and automation. A prototype architecture for the chorus network is also presented.

\section{The RobotChain}
\subsection{Previous Work}
\begin{figure}[t]
        \centering
        \includegraphics[width=8cm]{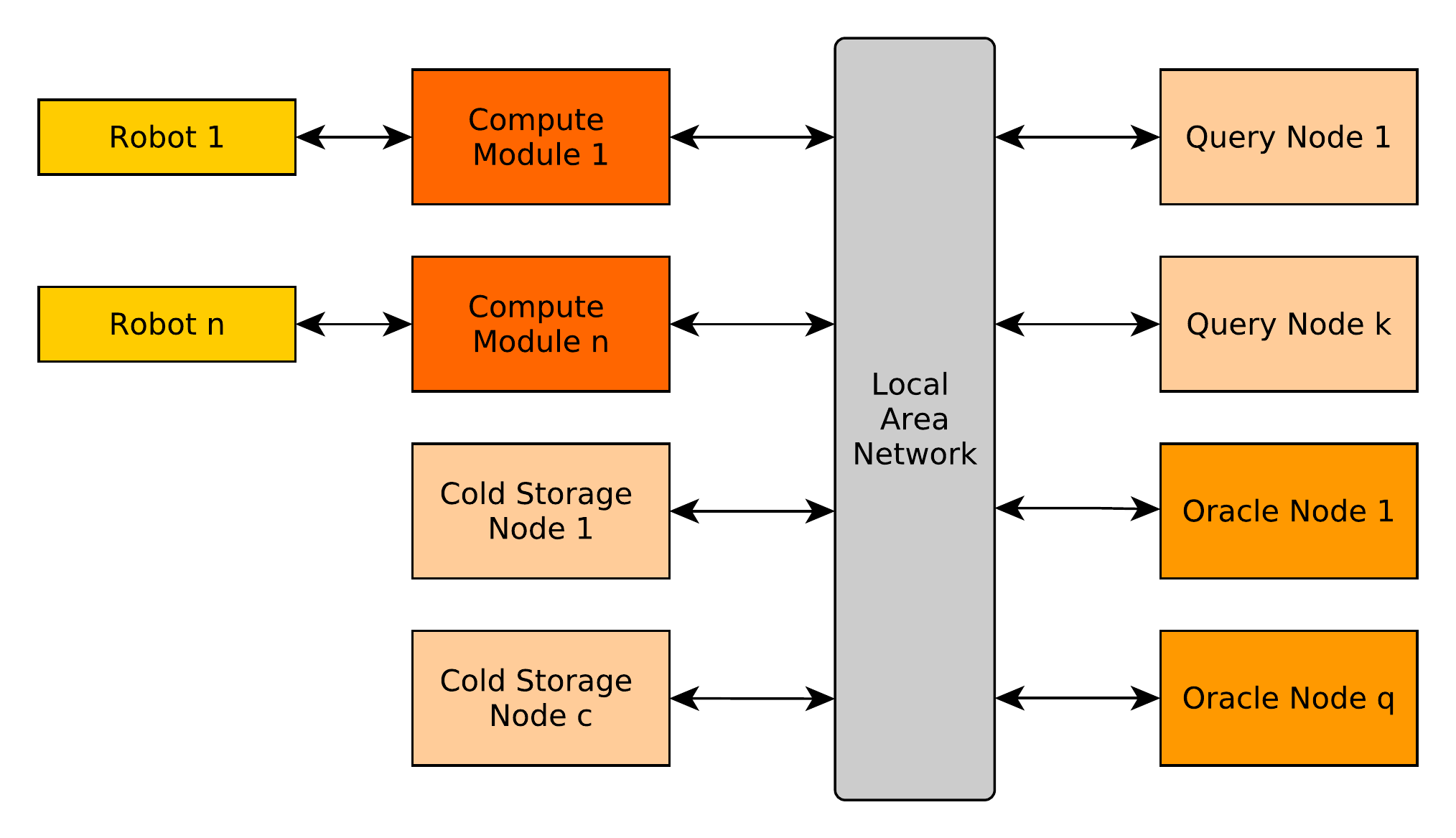}
        \caption{Overview of the RobotChain. The compute modules are devices that serve as interface between robots and RobotChain, Cold Storage nodes will save all the segments of the blockchain, Query nodes allow queries to the blockchain and Oracles are external entities that interact with the blockchain through smart contracts. One Cold Storage will also serve as the Genesis Node (see main text for details). }
        \label{fig:sys}
\end{figure}

%TODO: AS IS FROM MIT

Goodman\cite{tezos2018} presents a self-amending crypto-ledger implemented in \textit{OCaml} called Tezos.
Instead of using a genesis block or hash, it starts with a genesis protocol, containing a genesis block similar to other blockchains, but with functions that allow the amending of the protocol such that it can evolve.

The main feature of this blockchain is the fact that it implements a protocol that can adapt itself by transforming a Context object.
These amendments work over cycles, these cycles take about three months and are suggested by a submission to the chain, and stakeholders may vote for these amendments.
These amendments, if accepted, are first inserted into a \textit{testnet}, and after that, a second confirming vote is made. If the second vote is successful, the amendments are  integrated into the main protocol.

These amendments are considered a positive point due to the fact that this allows the community to enact changes in the blockchain, in order to improve it, similar to a political system, preventing blockchain hard forks, which are a radical change that results in divergence from the already created chain.

In addition to this, the fact that Tezos is open source allows us to adjust it to our proposes, mainly by trying to improve transaction speed, and possibly reduce complex steps in order to make it work on cheap computational units.
Another of the positive points of using Tezos blockchain technology is an increase of security with respect to the manipulation of the ledger. Also, smart contracts are proven correct, giving an additional layer of trust on the way the system is implemented.
In comparison to other mainstream protocols, the self-amending feature present in the Tezos network and the verifiable security using the OCaml language, are the distinguishing points provided by this network.
The self-amending feature provides a way to upgrade blockchain network, such as fine tuning the consensus algorithm, or other algorithms present without creating a hard fork on the already existent network, and the fact that the code is proved correct is of extreme importance when dealing with high cost equipment that can condition the operation of a factory.

%Modern factory assembly lines, are mainly operated by robotic units that require maintenance in order to operate at its peak performance. 
Robots are becoming ever more important in modern factories.
Currently it is a difficult task to keep track of every single action performed by every robot, in order to understand where possible bottlenecks are present or which robots need tuning, maintenance or even replacement. 
%Since maintenance of these operating units is a costly event, malicious entities may attempt to modify existing operation logs in order to either mask or cast blame onto other robots.
RobotChain contemplates the use of blockchain technology in order to solve the problem of keeping accurate immutable records of robotic actions in a factory environment. A public access blockchain is not desired, since the factory environment is a private environment and, as such, management does not allow outside access to its internal manufacturing information so RobotChain is a consortium blockchain: it has many of the advantages of a private blockchain, but instead of a single entity being the leader it operates under the leadership of a group, to allow for trust to be developed among the factory owners and the equipment providers.

Due to the fact that we are dealing with a set of robots from multiple manufacturers that are working at the same factor. We need that all the robot manufacturers and also the factory management, trust the event records, in the case that there is an accident and the guilty part has to be determined. The event records, that are stored in the blockchain, can be used for further goals such as understanding and improving manufacturing productivity, monitoring robots for malfunctioning\cite{8374192} or ensuring safety around working robots\cite{3dvision}. This paper deals with the fact that, although these records are important for the managers of the factories, they are not important for the day-to-day processing of the robot, since what a robot did four months ago is not important to the current functioning of the robot, meaning that is not needed on the limited storage of the compute device. So, this paper improves upon the original proposal of RobotChain\cite{robotchaingenesis}, with the introduction of the time-segmentation proposal.

RobotChain as already served as basis for controlling and monitoring robots. In \cite{8374192}, the authors show how it is possible to use RobotChain to register events and with the use of smart-contracts, detect when robots are having abnormal behaviours, which can indicate the presence of internal problems to the robot (faulty bearings or others) and external problems to it, such as problems in a production line. 

RobotChain was also shown to be capable of serving as basis to a new way of controlling robots \cite{2019arXiv190300660L}. In this, smart-contracts are used to store information about the analytics that Oracles (external entities that interact with the blockchain through smart contracts) acquire from images, which is used to detect how much material is waiting to be picked, and then, the smart-contracts are also used to control the robot velocity so it adjusts to the velocity of the production line (or even stops if there is no material to be picked). 
Over this scenario of picking materials in a production line, the authors of \cite{3dvision} show how multiple Oracles and multiple smart-contracts can be used to both monitor a robot workspace using 3D images and to control the robot to adjust its speed to the entrance of people in its workspace. 
In that work, the authors define two areas over the robot workspace, the warning zone and the critical zone, which are used to define the robot speed depending if the person that enters the different zones is known or unknown. If an unknown person enters the warning zone, a smart-contract defines that the velocity of the robot should be slowed down and if the person enters the critical zone, the robot is stopped. If the person is known and enters the warning zone, nothing happens but on entrance on the critical zone, the robot is slowed.
%VASCO TODO:

%- Our proposal (3.1 - The goals) -> What is the current solution for the problem you describe? How do factories ... "track every single action performed by every robot, in order to understand where possible bottlenecks are present?" Present this and defend your ideas in comparison to the current standards.
%\color{green}
%Our proposal contemplates the use of blockchain technology in order to solve the problem related to keeping accurate immutable records of robotic actions in a factory environment. Due to the intimate connection with this, a private blockchain network is required, in order to prevent outside unwanted access to the system, avoiding the addition of malware entry points. 
%\color{black}

\subsection{The Time-Segmentation Idea}
Figure \ref{fig:sys} presents RobotChain in a schematic way. Each robot is connected to a computation module, and this connection is bidirectional in order to receive information from the robot to feed the blockchain and allow the blockchain, via smart contracts, to change the robot behaviour.
%compute module é para tornar uniforme o accesso dos robos, seja uniformizar a informação que estes poem na rede, ou seja interface que estes utilizem.
The use of computational modules serves to ensure a uniform input into the blockchain, as different robots may need different connection interfaces. It also ensures that the robots are not negatively affected with additional software running, that could cause degraded performance or other unforeseen consequences.
In addition, there can be query nodes connected to the blockchain in order to query it for information. These are important to understand possible production line bottlenecks, or improve management understanding of the factory without directly interfacing with the robotic units.
The main concern is the high transaction volume that the various networked robots will produce. Also important is the fact that RobotChain may not impact in any form the performance of the robots.
% \color{black}
% Oracle nodes, as mentioned on the previous subsection, are external entities that interact with the blockchain by the means of smart contracts.
% \color{black}
Due to the fact that our proposal uses compute devices instead of running its code directly on the robot in order to prevent robot performance loss, the embedded compute devices are limited regarding data storage, as such, it is unfeasible to maintain a copy of the entirety of the blockchain on each device in the long term. As such, other nodes can be added to the network with the sole propose of cold storage.
% \color{black}
Although the proposed time segmentation is applied to a Tezos blockchain network running RobotChain, the modifications made are protocol agnostic, meaning that it can run under any protocol that Tezos supports.
% \color{black}

\subsection{Tezos History Mode}
%https://blog.nomadic-labs.com/introducing-snapshots-and-history-modes-for-the-tezos-node.html
As of 4 of February of 2019, Tezos has a new feature\cite{nomadiclabsblog} similar to the proposed solution, the history mode. This history mode changes how a node keeps its past data, with three different modes.
These modes rely on the checkpoint feature present on Tezos blockchain, where these checkpoints act as a regular interval anchor of consensus.

These modes are the archive mode, where the node keeps everything, which corresponds to the current Tezos default working mode, the full mode that is going to be the new default, where the node stores all data from the beginning of the chain, but drops information from previous checkpoints, keeping the headers and the operations from the previous checkpoints.
The last mode is the rolling mode, where the nodes only keep the latest checkpoint, effectively deleting old information.\label{sec:rollingmode}
These modes are a configuration parameter of a node in the Tezos blockchain.

As stated by the developers, this new feature has restrictions between mode changing, such as upgrading a node from full to archive.

Initialisation of the nodes, regardless of the activated mode, is still based on the regular blockchain synchronisation method via peer-to-peer or with the new snapshot feature, consisting on a file import/export.
This import feature does the same checks as synchronising from other nodes, recomputing and checking all the hashes encountered in the snapshot, effectively doing the same work as the peer-to-peer synchronisation but without the network cost.
This import file does not provide any security features regarding that the imported information from the file is equal to the information from the current main chain in the Tezos Network, meaning that the imported file might not be equal to Tezos main network but still be valid, possibly it may be maliciously modified, or belong to a private Tezos Network. Manual verification of the imported information, such as checking if the imported hash is contained on Tezos main network is needed.
These export snapshots must be created from full or archive nodes.
Comparison to our proposed method is presented in subsection \ref{subsec:comparisson_tezos_out}.

\section{Time-Segmentation Implementation}
\subsection{Overview}
\begin{figure*}[t]
        \centering
        \includegraphics[width=\textwidth]{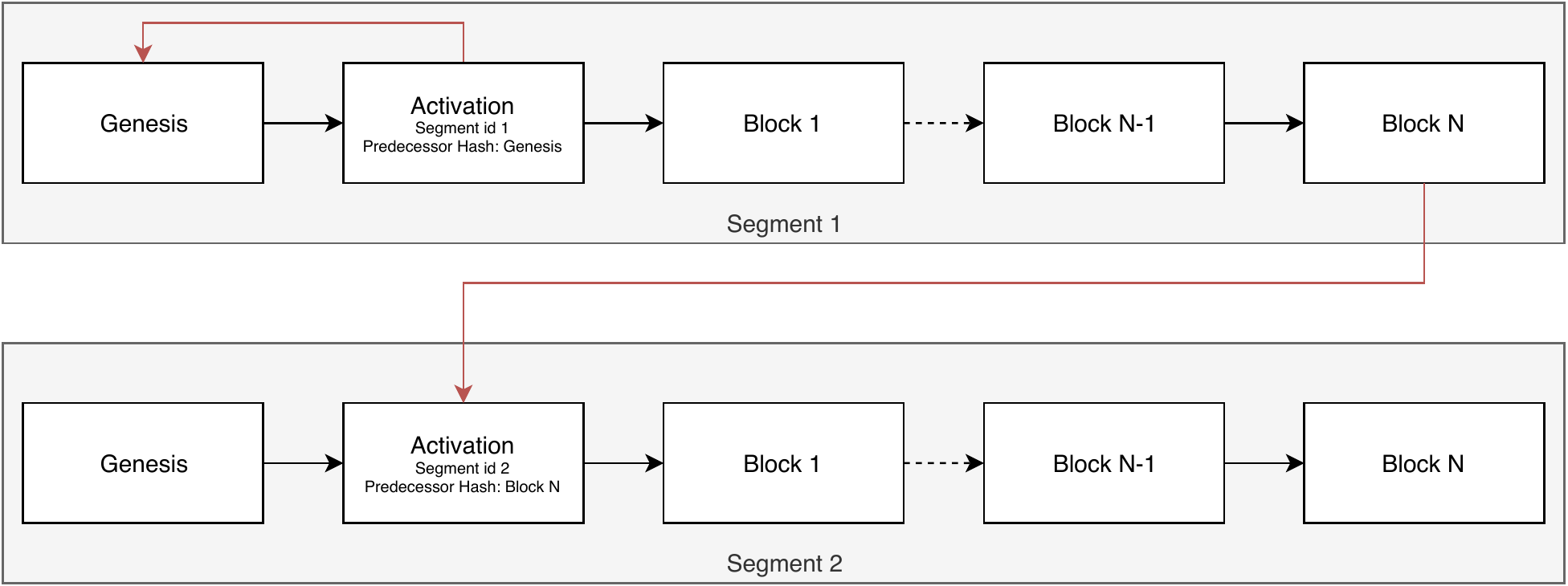}
        \caption{Visual representation of the approach of the time-segmentation process. The red arrows represent the use of hashes to allow inter-segment connectivity. Each activation block has the current segment id and the hash of the last block of the previous segment. In the case of the first segment, the predecessor hash is considered the genesis block hash, making segment 1 the "genesis segment".}
        \label{fig:segschm}
\end{figure*}
As mentioned in the previous section, this proposal presents a time-segmentation solution for consortium blockchains, implemented using Tezos blockchain technology. %\color{red}providing storage tests with relation to number of blocks generated in the network.\color{black}
This solution of time-segmentation of a blockchain consists in creating linked sub-blockchains, referred to as segment or segments throughout the paper, allowing compute devices with low storage capacity the ability to keep only the latest segment instead of the entire blockchain, while maintaining the non-modification of the chain itself. Non-modification is ensured via the first block on the new segment, in our case, the RobotChain protocol activation, containing the segment identifier (an integer) and the hash of the last block of the previous segment. A second block is also created in order to re-insert the various smart contracts present on the previous segment.

On the network, three types of nodes are now introduced: the genesis node, the cold storage nodes and the compute device nodes, where the genesis is meant to serve as a bootstrap point, protocol activation and smart contract initialisation and cold storage of the previous segments. The cold storage type is meant to only store all the segments and to retrieve older segments as needed and aid the bootstrap process, and the compute devices nodes are the interface between the blockchain and the robots as mentioned in the previous section. This solution is presented to solve the storage limitation of the compute devices and the fact that the older blocks are not entirely relevant to the continuous processing of the robot in a factory floor or the current state of the blockchain.
This allows compute devices with possibly small storage capacity in relation to regular computer hard drives to support a blockchain solution for an arbitrarily long time period. The three types of nodes (cold storage node, genesis node and compute device node) are running in the archive mode that is presented in subsection \ref{sec:rollingmode}, since the modifications are enacted on the Tezos blockchain version with the history mode.
Clearly, the number of cold storage nodes is typically much smaller than the number of compute device nodes.

%In this proof of concept that is still in development, the blockchain creates each segment with connection to the previous segment, several features under development are referred in the future work section, such as a RPC interface for cold storage block retrieval or the implementation of smart contracts on the network.
% \color{red}the blockchain only prunes the transaction records of the previous segments, maintaining a chained link to the genesis block allowing a improvement of the storage needed to keep the latest segment on the network, while providing the immutability of the blockchain technology.\color{black}

\subsection{Segment Creation Process}
% Two approaches were devised and tested during the development this proposal. 

% The first attempt works as follows.
% On the $N$-th block, the blockchain creates a copy of its current state, effectively backing it up, then it trims the information of the existent blocks, removing the information regarding transactions from the previous segments, and inserts the new block onto the blockchain, untrimmed, starting the new segment.
% Cold storage nodes would keep all the various segments and the compute devices would only keep the latest segment.

% Node synchronisation problems with this approach and the fact that we could not assert that the occupied storage had a maximum value, deemed it unreliable and development was focused on the second approach.

% Regarding the second approach, it works as follows:

The first segment, segment 1, starts the network as a regular blockchain, with the activation block receiving as parameters the segment id 1 and the original genesis block hash. Then, on its $N$-th block, the blockchain increments the segment ID on the node's configuration file and shuts down the validation and database parts of the blockchain, leaving the peer-to-peer interface online. By leaving the peer-to-peer interface online, there is no need to re-initialise the peer-to-peer interface or to rediscover peers. 

The state and validation are then reactivated with the updated configuration file, creating a new segment from scratch. The genesis node then activates the protocol, receiving as parameter the current segment ID and the predecessor segment hash, the hash from the last block of the previous segment, with the other nodes receiving this activation and resuming normal operations. 

Block one is then used to initialise smart contracts present on the previous segment and other features needed. Figure \ref{fig:segschm} presents this process in a visual way.
Considering a new compute device node that joins the blockchain, it will only synchronise the latest segment, the one running on the network currently.
In the case that the segmentation process happens before the bootstrap finishes, the node receives a reboot signal sent by the genesis node that instructs the node to create the new segment as described previously.

In the case that the new node is a cold storage node, the node will synchronise previous segments and keep the previous segments saved instead of deleting them.

\label{subsec:comparisson_tezos_out} % mode , without using the checkpoint system ,where
The presented approach is similar to the new Tezos history mode, where the compute device nodes would correspond to Tezos nodes running in the rolling mode presented in section \ref{sec:rollingmode}, and the cold storage nodes would work as the archive mode.
There are several differences between our proposal and the Tezos history mode such as node initialisation, where the the node has to synchronise from the genesis block up to the current information, or use the snapshot feature introduced.
In the case of factory fast pace environment, creating snapshots and waiting for the new elements to catch-up in order to start operations may not be feasible.

Since the created approach is based on the Tezos history mode, it may be configured to work with the presented three modes, the archive, full and rolling modes, and, as such, the archive was the selected mode for running in both the cold storage nodes and the compute device nodes.

Our solution has the advantage of only needing to synchronise the latest segment, without the need to request older segment information, with the sole exception of the activation node's previous segment hash that is needed for protocol activation only once per segment.
This advantage counterpoints the need of synchronising the entirety of the blockchain data from a snapshot generated from a full/archive node.

%Later on, cold storage nodes will have the ability to request segments to other cold storage nodes.
% Before Every Nth block is written into storage, a copy of the database file is created, followed by a prune of the transactions contained in each preceding block, with the exception of the genesis block and the first block that matches the protocol activation. Then the current block is inserted with complete information and the process repeats in the next Nth block.

% New nodes inserted into the network currently receive every block in the blockchain already pruned with the current block segment without pruning.(re verificar)
\section{Experimental Results}
\subsection{Compute Device Node Storage}

% \begin{figure}[t]
%         \centering
%         \includegraphics[width=8cm]{figs/modes_vs_seg.pdf}
%         \caption{\textcolor{black}{RECALCULAR RESULTADOS. Storage size on each regular node (not cold storage), for four versions of the blockchain, as a function of the number of blocks. For the segmented blockchain (normal), the running mode considered is archive.}}
%         \label{fig:results}
% \end{figure}
\begin{figure}[t]
        \centering
        \subfloat{\includegraphics[width=.46\linewidth]{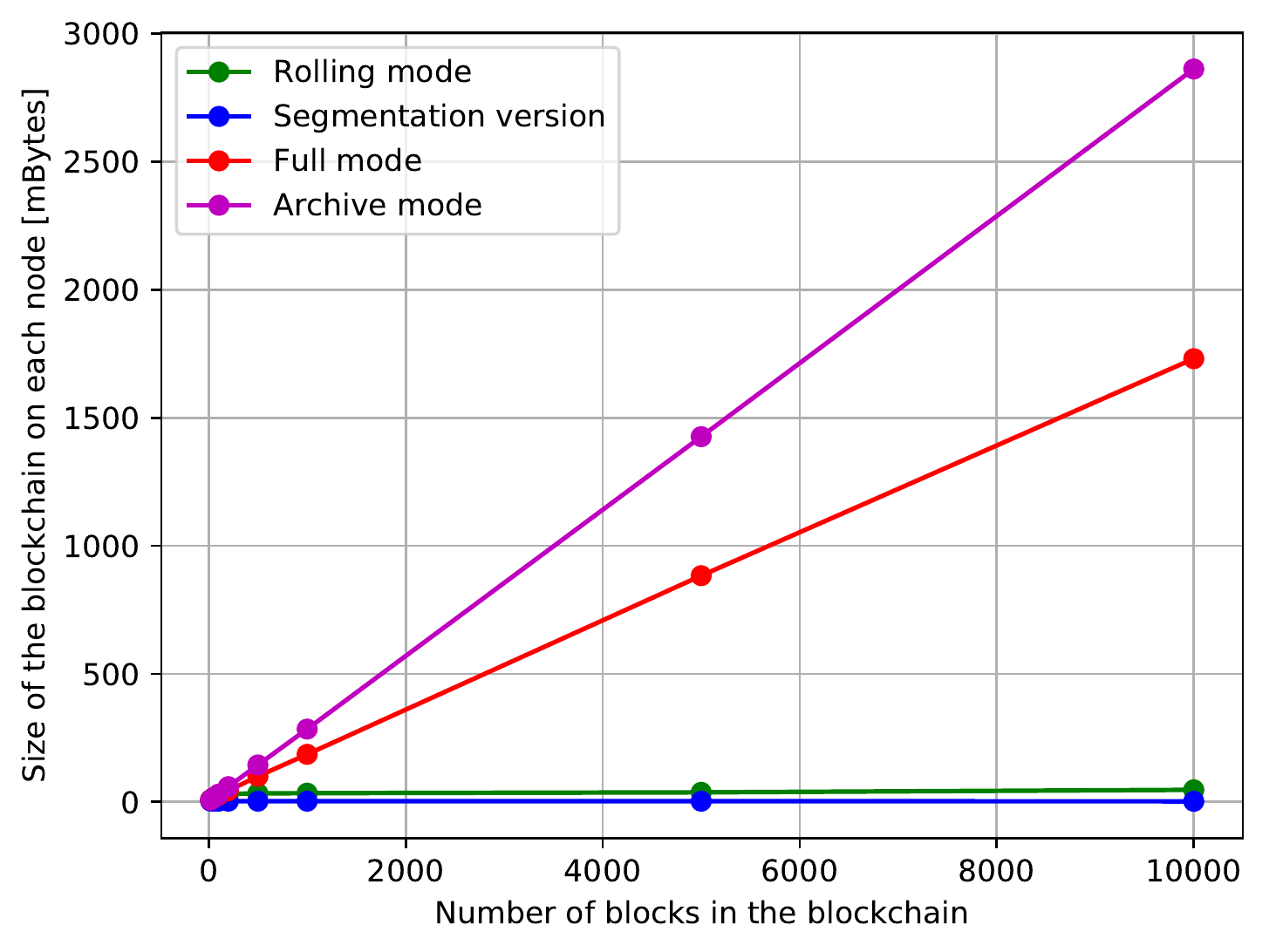}}
        \qquad
        \subfloat{\includegraphics[width=.46\linewidth]{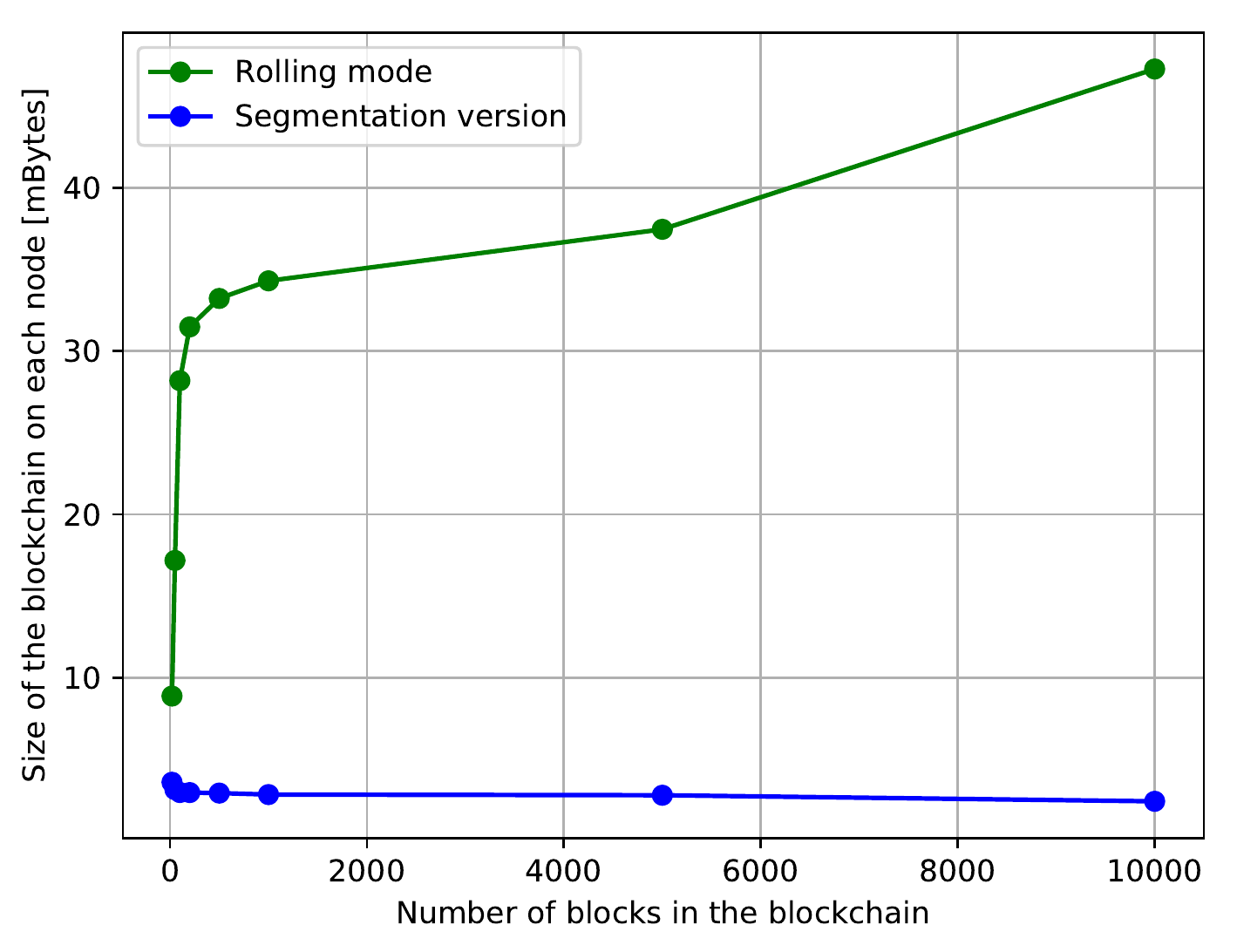}}
        \caption{Storage size on each compute device node (not cold storage), for four versions of the blockchain, as a function of the number of blocks. The segmented blockchain running mode is archive. The left sub-figure presents the results for the four tests, while the right sub-figure presents only the results for the rolling mode and the segmentation version, for improved comparison.}
        \label{fig:results}
\end{figure}

Tests were made to evaluate the storage requirements of our proposal, comparing the unsegmented blockchain, running the three versions (archive, full, rolling) and the approach configured as full, to build the segmented blockchain, segmenting the blockchain every 10 blocks.
The tests were ran varying the total number of blocks. Random transactions were injected into the network up to a maximum of 32 transaction clients running at the same time. These clients inject a transaction between random accounts, with value 1, with a transaction description of a minimum 1000 random characters.
The storage test results are provided in figure \ref{fig:results}.

The presented approach has a average size of 2282 Kilobytes per segment, having a definite hard limit for the maximum storage occupied by each segment, using 10 as the number of blocks per segment. The fact that our proposal fixes a maximum size per segment allows the blockchain to increase its longevity arbitrarily. It also aids with bootstrap due to the fact that a new node won't need to obtain every segment from the start, needing only the latest one to work.

With the ability of creating segments limited with respect to the necessary storage space, RAM disk execution on the compute devices is a possibility under investigation in order to improve the blockchain speed.

\subsection{Cold Storage Node}

% \begin{figure}[t]
%         \centering
%         \includegraphics[width=8cm]{figs/fig4/corrected.pdf}
%         \caption{\textcolor{black}{Storage size on cold storage nodes and compute device nodes for the segmentation approach blockchain, as a function of the number of blocks in the segment. The compute device size is the size of a segment in MBytes. The cold storage size is the storage occupied by all the segments created in MBytes in a cold storage node. For comparison, each running mode storage space occupied by the unsegmented network in MBytes is presented.}}
%         \label{fig:results_cs}
% \end{figure}

\begin{figure}[t]
        \centering
        \subfloat{\includegraphics[width=.46\linewidth]{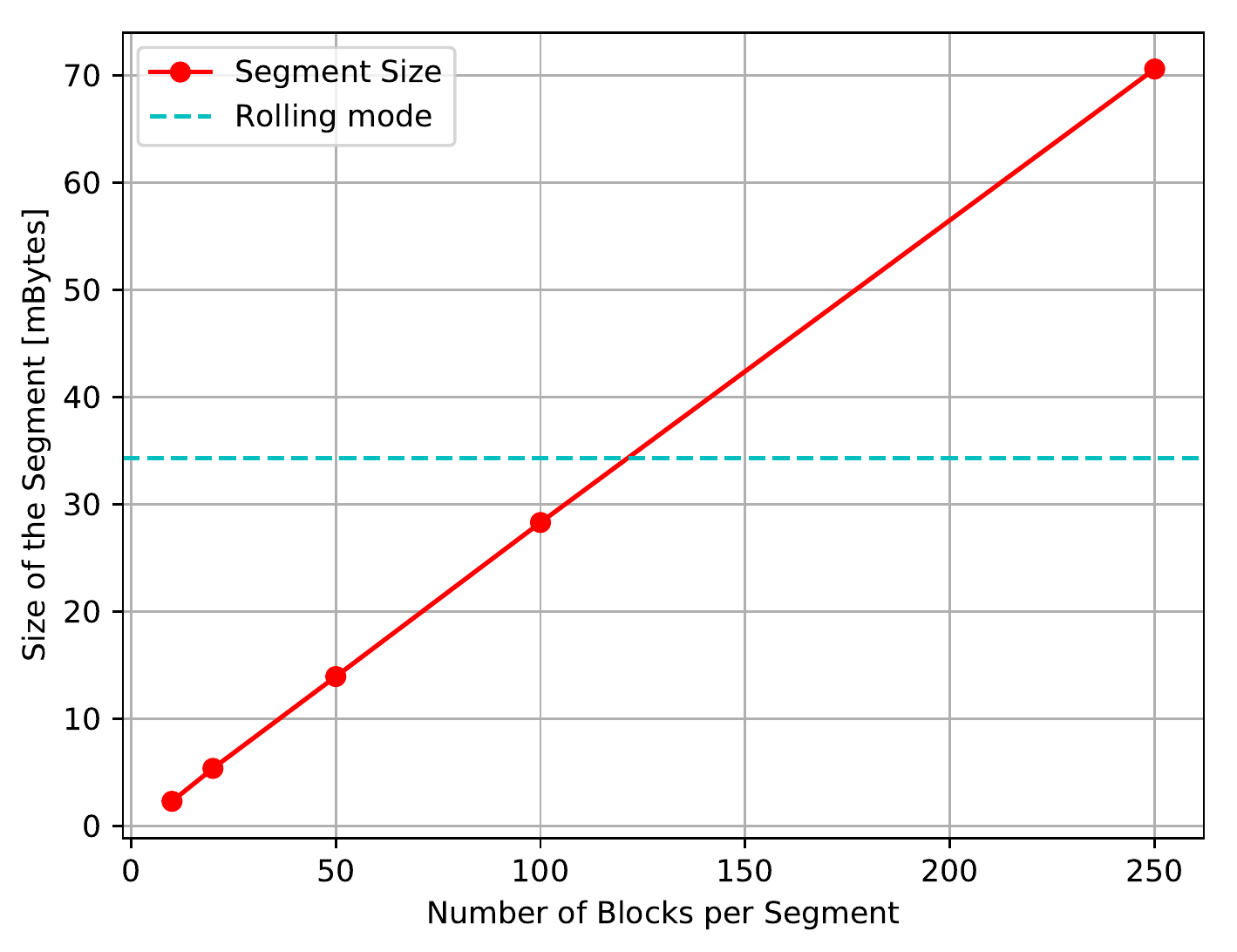}}
        \qquad
        \subfloat{\includegraphics[width=.46\linewidth]{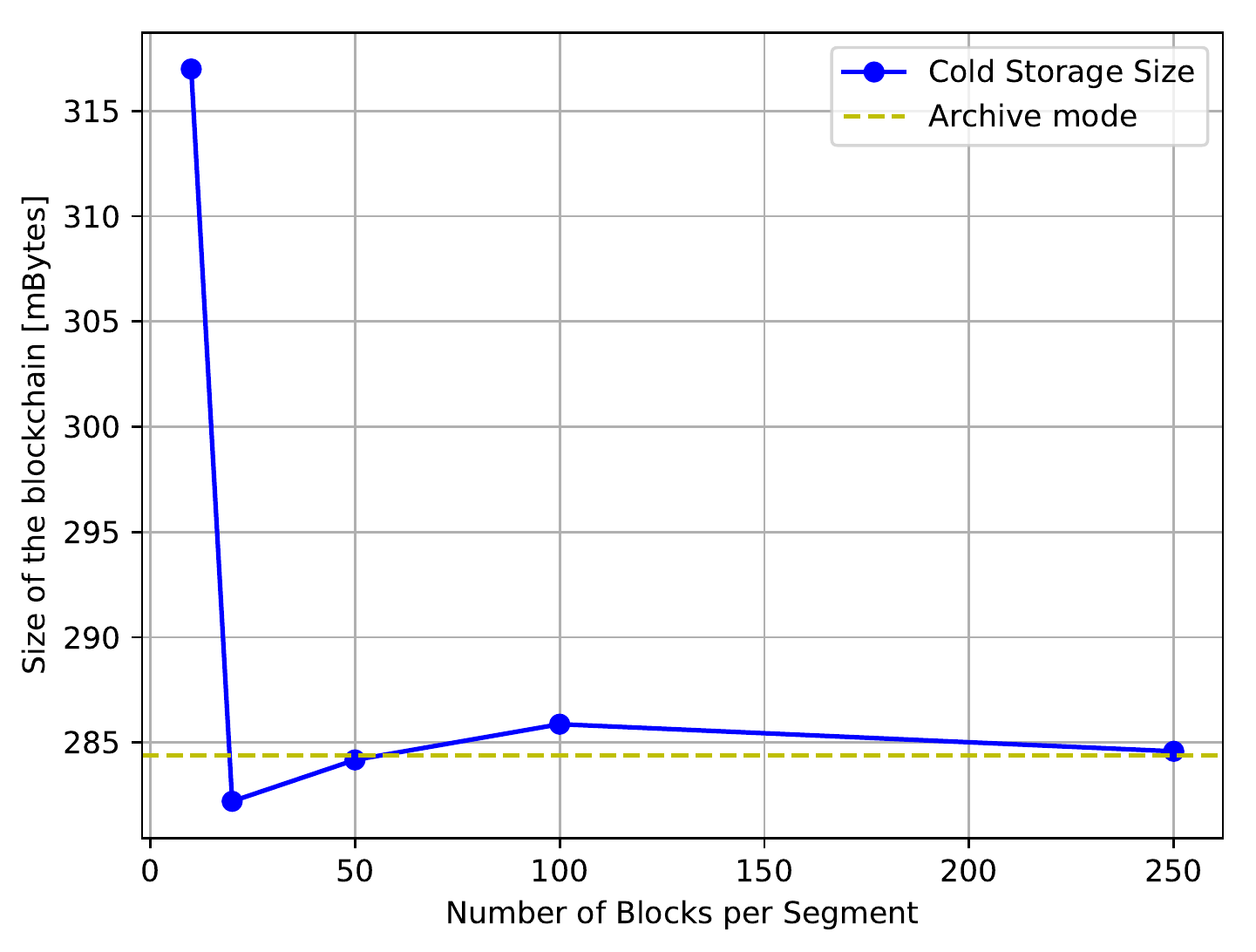}}
        \caption{The left sub-figure presents the compute device storage size as a function of the number of blocks per segment for a total of 1000 blocks, compared with the corresponding rolling mode storage size of the Tezos blockchain. The right sub-figure presents the cold storage size as a number of blocks per segment for a total of 1000 blocks, compared with the archive modes of the Tezos blockchain. Results for the full mode are not presented since they are not comparable to any of the node types on our approach, regarding storage capabilities.}
        \label{fig:results_cs}
\end{figure}

Additional tests were made to understand how the segmentation affected the storage capabilities of both the compute device nodes and the cold storage nodes, and how the network would grow with the various segments, considering the repeated addition of the genesis and activation blocks to each segment.
The tests were ran for 20, 50, 100, 250 blocks per segment with a total of 1000 blocks per experiment. As with the previous experiment, random transactions were injected into the network up to a maximum of 32 transaction clients running at the same time. These clients inject a transaction between random accounts, with a value of 1, with a transaction description of a minimum 1000 random characters. The storage results are presented in Fig. \ref{fig:results_cs}.

The segmentation approach has a smaller running storage footprint, with a cold storage space occupied similar to archive node.
Depending on the value for the number of segments per block, the resulting segment storage space occupied can be inferior to the rolling mode.
%Nonetheless we can see that choosing an appropriate value for the size of the segments can keep them arbitrarily small with the global size penalisation of about 10 to 20\% on the cold storage nodes, when compared to what would happen with the non-segmented blockchain. Note also that, as the number of blocks in a segment increases, the global size penalisation decreases (black line in Fig.4 converges to the green line). Anyway, we do not see a problem in requiring a penalisation of global size of around 20\% on the cold storage nodes (which will be a small number of nodes) when compared to the benefits of having fast and low cost nodes running the blockchain, even in their RAMs.
As referred, our approach as the benefit of providing a definite hard limit for the segments, with a cold storage space occupied similar to a node running archive mode, considering the increased number of blocks with the added genesis and activation block. This allows low cost nodes running the blockchain, with an increase speed with the possibility of running in a RAM disk.
In addition, defining an appropriate segment value like $100$, it can also have a smaller size per segment when compared to the rolling mode. There is also the advantage of not needing to bootstrap the entire network and just needing the latest segment.

Finally it is important to state that the both the cold storage node and the compute device node are both running in archive mode and that Tezos rolling mode does not guarantee a fixed maximum node memory size, and the memory requirements slowly grow as can be seen in Fig.\ref{fig:results} (right). This invalidates the Tezos rolling mode as a solution to the limited capacity of the said nodes, since the rolling mode would eventually exhaust the available memory and the network would stop working.

\section{Conclusions}
This paper improves upon the original proposal of RobotChain\cite{robotchaingenesis}, a robotic event storage solution, that enables robot monitoring, control and cooperation, with the introduction of the time-segmentation proposal to solve the problems related to the small storage capacity of compute modules.
It allows the use of cheap compute modules for the majority of network nodes (all but the cold storage ones) and makes the processing and connection of new nodes faster both by allowing the use of faster memory for storing the segment (such as RAM) and also because only the current segment is needed for syncing the new node with the network.
The solution presented allows the creation of a time-segmented blockchain that has a definite hard limit on the segments' storage capacity, that is independent of how long the blockchain has run for.

The new history feature from Tezos, although similar to the proposed method, has differences and requirements that deemed it not compatible with the performance that RobotChain needs to achieve. These differences include the time needed to include a new node in the RobotChain, the information needed for the robots to operate and the capability of defining a storage space hard limit that allows deployment of the compute modules for an unlimited duration.

% \section{Future work}

As future work, several features related to this time-segmentation solution are to be implemented such as RPC interfaces for block retrieval for previous segments, allowing query nodes to access information not contained on the current segment or cold storage full synchronisation by allowing new cold storage nodes to obtain previous segments from other cold storage nodes.
As an alternative, the snapshot feature will also be investigated for the purposes of cold storage node bootstrap. We also intend to study the possibility of defining a different number of blocks per segment on a on node-to-node basis, that could be useful for accommodating nodes with different capabilities.
%\color{black}



%% References


\bibliographystyle{model1-num-names}
\bibliography{root.bib}

\begin{thebibliography}{12}
\expandafter\ifx\csname natexlab\endcsname\relax\def\natexlab#1{#1}\fi
\providecommand{\bibinfo}[2]{#2}
\ifx\xfnm\relax \def\xfnm[#1]{\unskip,\space#1}\fi
%Type = Article
\bibitem[{Ferrer(2016)}]{Ferrer2016}
\bibinfo{author}{E.~C. Ferrer},
\newblock \bibinfo{title}{{The blockchain: a new framework for robotic swarm
  systems}}  (\bibinfo{year}{2016}).
%Type = Techreport
\bibitem[{Strobel et~al.(2018)Strobel, {Castello Ferrer}, and
  Dorigo}]{Strobel2018}
\bibinfo{author}{V.~Strobel}, \bibinfo{author}{E.~{Castello Ferrer}},
  \bibinfo{author}{M.~Dorigo}, \bibinfo{title}{{Managing Byzantine Robots via
  Blockchain Technology in a Swarm Robotics Collective Decision Making
  Scenario}}, \bibinfo{type}{Technical Report}, \bibinfo{year}{2018}.
%Type = Article
\bibitem[{Danilov et~al.(2018)Danilov, Rezin, Kolotov, and
  Afanasyev}]{Danilov2018}
\bibinfo{author}{K.~Danilov}, \bibinfo{author}{R.~Rezin},
  \bibinfo{author}{A.~Kolotov}, \bibinfo{author}{I.~Afanasyev},
\newblock \bibinfo{title}{{Towards blockchain-based robonomics: autonomous
  agents behavior validation}}  (\bibinfo{year}{2018}).
%Type = Article
\bibitem[{Ferrer et~al.(2018)Ferrer, Rudovic, Hardjono, and
  Pentland}]{Ferrer2018}
\bibinfo{author}{E.~C. Ferrer}, \bibinfo{author}{O.~Rudovic},
  \bibinfo{author}{T.~Hardjono}, \bibinfo{author}{A.~Pentland},
\newblock \bibinfo{title}{{RoboChain: A Secure Data-Sharing Framework for
  Human-Robot Interaction}}  (\bibinfo{year}{2018}).
%Type = Inproceedings
\bibitem[{Hasan et~al.(2018)Hasan, Datta, Rahman, and Shahriar}]{8377911}
\bibinfo{author}{M.~G. M.~M. Hasan}, \bibinfo{author}{A.~Datta},
  \bibinfo{author}{M.~A. Rahman}, \bibinfo{author}{H.~Shahriar},
\newblock \bibinfo{title}{Chained of things: A secure and dependable design of
  autonomous vehicle services},
\newblock in: \bibinfo{booktitle}{2018 IEEE 42nd Annual Computer Software and
  Applications Conference (COMPSAC)}, volume~\bibinfo{volume}{02}, pp.
  \bibinfo{pages}{498--503}.
%Type = Article
\bibitem[{Leiding and Vorobev(2018)}]{chorus2018}
\bibinfo{author}{B.~Leiding}, \bibinfo{author}{W.~V. Vorobev},
\newblock \bibinfo{title}{Enabling the vehicle economy using ablockchain-based
  value transaction layerprotocol for vehicular ad-hoc networks}
  (\bibinfo{year}{2018}).
%Type = Article
\bibitem[{Goodman(2008)}]{tezos2018}
\bibinfo{author}{L.~Goodman},
\newblock \bibinfo{title}{Tezos - a self-amending crypto-ledger}
  (\bibinfo{year}{2008}).
%Type = Inproceedings
\bibitem[{{Lopes} and {Alexandre}(2019)}]{8374192}
\bibinfo{author}{V.~{Lopes}}, \bibinfo{author}{L.~A. {Alexandre}},
\newblock \bibinfo{title}{Detecting robotic anomalies using robotchain},
\newblock in: \bibinfo{booktitle}{2019 IEEE International Conference on
  Autonomous Robot Systems and Competitions (ICARSC)}.
%Type = Article
\bibitem[{Lopes and Alexandre(2019)}]{3dvision}
\bibinfo{author}{V.~Lopes}, \bibinfo{author}{L.~A. Alexandre},
\newblock \bibinfo{title}{{Robot Workspace Monitoring using a Blockchain-based
  3D Vision Approach}},
\newblock \bibinfo{journal}{submitted}  (\bibinfo{year}{2019}).
%Type = Article
\bibitem[{{Fernandes} and {Alexandre}(2018)}]{robotchaingenesis}
\bibinfo{author}{M.~{Fernandes}}, \bibinfo{author}{L.~A. {Alexandre}},
\newblock \bibinfo{title}{{Robotchain: Using Tezos Technology for Robot Event
  Management}}  (\bibinfo{year}{2018}).
%Type = Article
\bibitem[{{Lopes} et~al.(2019){Lopes}, {Alexandre}, and
  {Pereira}}]{2019arXiv190300660L}
\bibinfo{author}{V.~{Lopes}}, \bibinfo{author}{L.~A. {Alexandre}},
  \bibinfo{author}{N.~{Pereira}},
\newblock \bibinfo{title}{{Controlling Robots using Artificial Intelligence and
  a Consortium Blockchain}}  (\bibinfo{year}{2019})
  \bibinfo{pages}{arXiv:1903.00660}.
%Type = Misc
\bibitem[{{Nomadic Labs}(2019)}]{nomadiclabsblog}
\bibinfo{author}{{Nomadic Labs}}, \bibinfo{title}{Introducing snapshots and
  history modes for the tezos node}, \bibinfo{year}{2019}.

\end{thebibliography}






% \end{thebibliography}

\end{document}